\documentclass[10pt,twocolumn,letterpaper]{article}

\usepackage{iccv}
\usepackage{times}
\usepackage{epsfig}
\usepackage{graphicx}
\usepackage{amsmath}
\usepackage{amssymb}
\usepackage{multirow}
\usepackage[colorlinks=true, urlcolor=blue, linkcolor=red]{hyperref}
\usepackage[ruled,linesnumbered]{algorithm2e}
\usepackage[accsupp]{axessibility}  

\iccvfinalcopy 

\def\iccvPaperID{11416} 

\ificcvfinal\fi

\begin{document}

\title{Hierarchical Point-based Active Learning for Semi-supervised Point Cloud Semantic Segmentation}

\author{Zongyi Xu\textsuperscript{1\dag}\quad
Bo Yuan\textsuperscript{1\dag}\quad
Shanshan Zhao\textsuperscript{2\dag}\quad
Qianni Zhang\textsuperscript{3}\quad
Xinbo Gao\textsuperscript{1}\thanks{Corresponding author. \dag Equal contribution.}\\
\textsuperscript{1} Chongqing University of Posts and Telecommunications, China\\
\textsuperscript{2} {JD Explore Academy, China}\quad
\textsuperscript{3} {Queen Mary University of London, UK}\\
{\tt\small  \{xuzy, gaoxb\}@cqupt.edu.cn}\quad
{\tt\small s210201127@stu.cqupt.edu.cn}\\
{\tt\small sshan.zhao00@gmail.com}\quad
{\tt\small  qianni.zhang@qmul.ac.uk}
}

\maketitle
\ificcvfinal\thispagestyle{empty}\fi


\begin{abstract}

 Impressive performance on point cloud semantic segmentation has been achieved by fully-supervised methods with large amounts of labelled data. As it is labour-intensive to acquire large-scale point cloud data with point-wise labels, many attempts have been made to explore learning 3D point cloud segmentation with limited annotations. 
Active learning is one of the effective strategies to achieve this purpose but is still under-explored. The most recent methods of this kind measure the uncertainty of each pre-divided region for manual labelling but they suffer from redundant information and require additional efforts for region division. 
This paper aims at addressing this issue by developing a hierarchical point-based active learning strategy. Specifically, we measure the uncertainty for each point by a hierarchical minimum margin uncertainty module which
considers the contextual information at multiple levels. Then, a feature-distance suppression strategy is designed to select important and representative points for manual labelling. 
Besides, to better exploit the unlabelled data, we build a semi-supervised segmentation framework based on our active strategy. Extensive experiments on the S3DIS and ScanNetV2 datasets demonstrate that the proposed framework achieves $96.5\%$ and $100\%$ performance of fully-supervised baseline with only $0.07\%$ and $0.1\%$ training data, respectively, outperforming the state-of-the-art weakly-supervised and active learning methods. The code will be available at \href{URL}{https://github.com/SmiletoE/HPAL}.
\end{abstract}

\section{Introduction}

Point cloud semantic segmentation aims to assign a category label for each 3D point, which can be applied to various scenarios, such as  robotics \cite{thomas2021self}, autonomous driving \cite{abbasi2022lidar}, and augmented reality \cite{lai2022stratified}. Recently, deep learning based methods \cite{wang2019graph, zhang2020polarnet} have achieved impressive performance. These high-performing methods usually rely on large amounts of data with point-wise labels. However, acquiring such dense labels for 3D point clouds is extremely tedious and costly.

 To relieve the labour and cost of annotation,  many recent methods explore semi-supervised learning to learn 3D segmentation models with limited labelled points that are usually sampled randomly. These methods attempt to employ effective strategies to propagate the label information to the unlabelled points~\cite{cheng2021sspc,ji2023semi,deng2022superpoint,hu2022sqn}. Although these semi-supervised methods can greatly decrease labelling costs, their performance might be limited due to various factors. Among them, the most important reason is that some of the randomly selected points are in fact redundant while some really important points might be omitted.

 As an alternative learning strategy, active learning recently is studied to alleviate such limitations for 3D segmentation.
Lin et al. \cite{lin2020efficient} divide the whole point clouds into segments and each segment is utilised as the basic query unit for sample selection.
 ReDAL~\cite{wu2021redal} proposes to select those informative and diverse sub-scene regions for label acquisition. The entropy, colour discontinuity, and structural complexity are used to measure the information of sub-scene regions.  Following it, SSDR-AL~\cite{shao2022active} groups the original point clouds into superpoints and incrementally selects the most informative regions for annotation.  
However, these methods usually rely on pre-dividing the point cloud into multiple regions, and the region division strategy might negatively impact the performance. Moreover, as point clouds present strong semantic similarity in local areas, selecting all the points in the local region results in a redundant labelling budget.  

Motivated by the above analysis, this research aims at enhancing the 3D segmentation performance with limited labelled data in the active learning framework. Specifically, in comparison with previous region-based methods, we measure the uncertainty or importance of labelling for each point to avoid the additional efforts and potential bias of region division. However, considering each point individually may introduce extra noise to the training process, leading to unreliable uncertainty values that cannot reflect the real importance of each point. 
To alleviate this issue, we design a hierarchical minimum margin uncertainty (HMMU) measurement module by considering the local contextual information. The HMMU involves the uncertainty of the point and its neighbourhood. In this way, the acquired score  can more effectively represent the importance of each point label.
Based on the uncertainty scores acquired by HMMU, we can directly choose the Top-$K$ points for labelling. However, 
as the point cloud presents semantic similarity in the local region, some redundant uncertain points tend to be distributed densely. To further reduce the labelling  cost, we propose a feature-distance  suppression module (FDS) to remove those uncertain points with similar features in the neighbourhood. By exploiting the devised HMMU measurement and FDS for redundant point filtering, our hierarchical point-based active learning strategy is able to select  more valuable points for labelling and thus reduce labelling costs.
Furthermore, inspired by previous semi-supervised methods that exploit the unlabelled points, we utilise a simple teacher-student scheme to enhance the supervision signal by assigning a pseudo label for the unlabelled points. 

The main contributions of this research are as follows:
\begin{itemize}
	\item We propose a point-based active learning method for semi-supervised point cloud semantic segmentation, which surpasses current semi-supervised or active learning
methods and achieves comparable performance with full supervision methods relying on scarce annotations.

	\item We propose a novel hierarchical minimum margin uncertainty module to measure the uncertainty for each point by progressively perceiving the contextual information at increasing scales.  

 \item We propose a feature-distance suppression module to  remove redundant points with similar features in the neighbourhood and further minimise the labelling cost.
 \end{itemize}
\begin{figure*}[t]
    \begin{center}
        \includegraphics[width=0.9\linewidth]{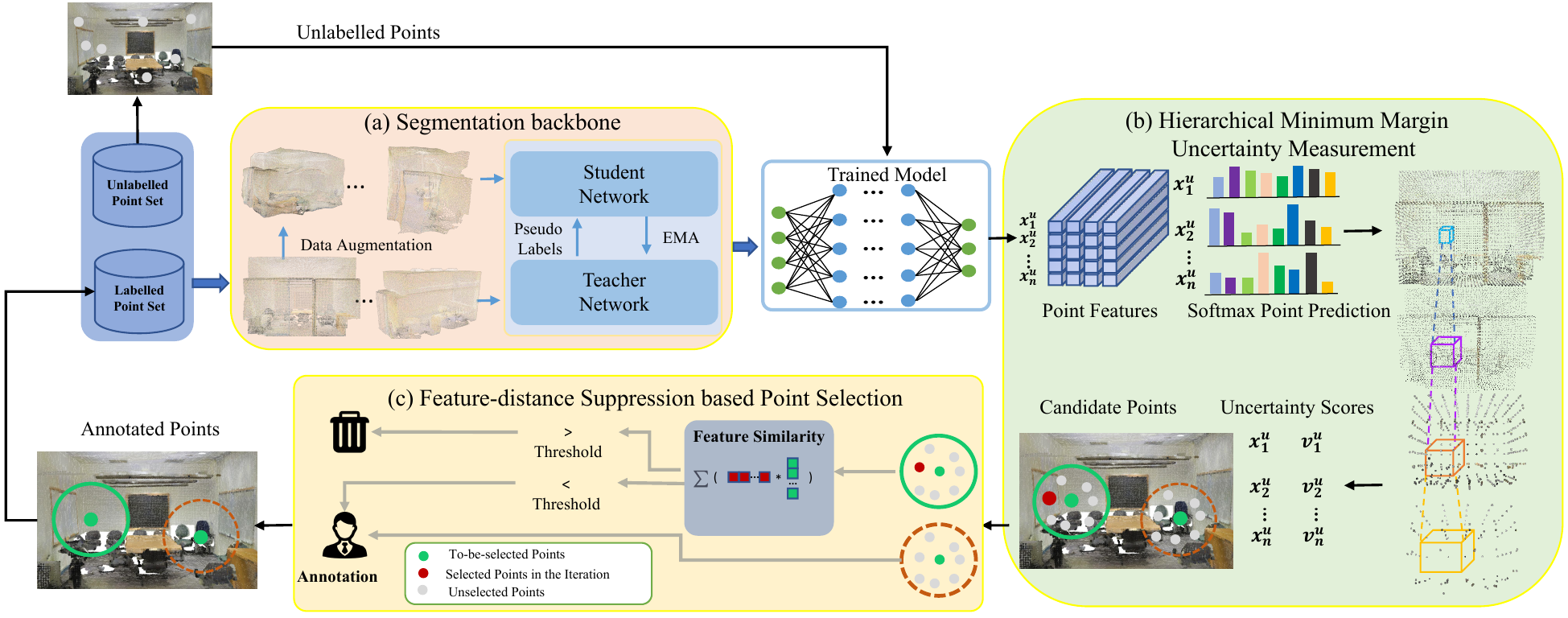}
    \end{center}
    \vspace{-1em}
   \caption{The proposed hierarchical point-based active learning framework for semi-supervised point cloud semantic segmentation. (a) A segmentation network using the teacher-student model is trained with the initial labelled and unlabelled training data.  The teacher is updated from the student by the exponential moving averaging (EMA) and the student network is obtained as our target segmentation model. (b) Those representative candidate points are measured by the trained model  using the proposed hierarchical minimum margin uncertainty module (HMMU).
   (c) Feature-distance Suppression module (FDS) is then applied to remove those redundant points in the neighbourhood and select the final to-be-annotated points. The selected valuable points are annotated and added to the labelled training set to improve the segmentation model.}
    \label{fig:framework}
    \vspace{-1em}
\end{figure*}
\section{Related Work}
\subsection{3D Semantic Segmentation}
In recent years, a large number of deep learning-based point cloud semantic segmentation methods have been developed, which can be roughly divided into the following categories:\textbf{ 1) 2D projection-based methods }~\cite{boulch2017unstructured,kundu2020virtual,aksoy2020salsanet,cortinhal2020salsanext,milioto2019rangenet++,wu2018squeezeseg,wu2019squeezesegv2,xu2020squeezesegv3}, which project the 3D point cloud onto 2D images through a variety of viewpoints including multi-view, bird's eye view, 
spherical projection etc. Projection-based methods can make full use of those well-designed 2D convolution networks for 3D scene parsing, but obviously, the projection causes the loss of 3D geometric information and thus limits the performance.
\textbf{ 2) voxel-based methods }~\cite{cheng20212,graham20183d,tang2020searching,yan2021sparse,choy20194d,tchapmi2017segcloud}. Point cloud voxelization regularizes point clouds with uneven densities, making it possible to extend regular 2D convolutions to 3D scenes. At an early stage, voxel-based methods usually suffer from
the explosion of computation and the loss of information. Fortunately, some sparse convolution methods~\cite{graham20183d,tang2020searching,choy20194d} have been proposed over time to greatly ease the computational costs.
\textbf{ 3) point-based methods}~\cite{hu2020randla,qi2017pointnet,qi2017pointnet++,thomas2019kpconv,li2018pointcnn,wu2019pointconv}. These methods directly take the original uneven point cloud as input and use MLPs, 3D point convolution, and other more flexible ways to process the point clouds, which preserve the original 3D information. However, their high performance still relies on a large number of point-level labels. 

\subsection{Weakly-supervised 3D Semantic Segmentation}
We collectively refer to all methods training with sparse annotation  as weakly supervised methods, which is a well-explored field in point cloud segmentation. These methods are usually based on a small set of randomly labelled data and then exploit more information from the unlabelled data by leveraging techniques such as transfer learning, contrastive learning, consistency regularization, pseudo-label, etc.
Based on the transfer learning technologies, pre-training is widely exploited in 2D images, and has recently been gradually applied to point clouds \cite{zhang2021weakly,yu2022data,xie2020pointcontrast,zhang2021self,wang2020weakly,xu2022image2point}. These methods learn pre-trained knowledge from 3D point clouds or 2D images and then apply it to the target dataset, thus achieving better training results with limited labelled data.
Some other works train the model by using consistency regularization \cite{xu2020weakly,zhang2021perturbed,li2022hybridcr,yang2022mil} and pseudo-label \cite{hu2022sqn,tian2022vibus,cheng2021sspc,liu2021one,zhang2021weakly,li2022hybridcr}. 
 SQN \cite{hu2022sqn} is one of these works which designs a novel semantic query network to efficiently obtain the pseudo-labels.
To further improve weakly supervised training, some approaches also use contrastive learning \cite{li2022hybridcr,liu2021one,hou2021exploring,jiang2021guided,xie2020pointcontrast}, which usually combine contrastive loss with pre-training, consistency regularization, and pseudo-labels to better exploit the supervision information. 
Most weakly-supervised methods explore information from the labelled data, so how to select the most informative data to label is a question of crucial importance.

\subsection{Active Learning on 3D Semantic Segmentation}
Similar to weakly-supervised learning, active learning also aims to train deep models with limited labelled data.
The main difference is that weakly-supervised learning mainly focuses on enhancing supervision by exploiting available label information while active learning focuses on the selection of valuable information for labelling.

Lin et al. \cite{lin2020efficient} make the first attempt for active learning on 3D semantic segmentation using segments as the basic query unit, and propose a segment entropy strategy to measure the informativeness of each segment.
To improve the efficiency of active learning, superpoints are introduced as the most common basic query units \cite{shi2021label,wu2021redal}. Considering the limitation of uncertainty, Shi et al. \cite{shi2021label} propose to design active acquisition functions from multiple perspectives, including feature diversity, shape diversity, and entropy. ReDAL \cite{wu2021redal} combines entropy with the geometric structure and colour information and uses feature clustering to prevent selection redundancy. In addition to the entropy-based method, SSDR-AL \cite{shao2022active} improves MMU based on super-point by considering the principal class within super-points.
To date, all active learning methods for point cloud semantic segmentation are region based. However, the pre-region division can greatly affect the active learning performance and unnecessary annotation in the semantic-similar areas is prone to occur. 

\section{Approach}
\subsection{Overview}
In the following, we describe our proposed hierarchical point-based active learning approach for semi-supervised point cloud semantic segmentation in detail.
As shown in Figure \ref{fig:framework}, the initial labelled  and   unlabelled training data are utilised to train the point cloud semantic segmentation neural network in a semi-supervised way. A teacher-student model is adopted. With the trained model, the valuable unlabelled points that are able to improve the segmentation performance most can be measured and selected in an active learning manner. The hierarchical minimum margin uncertainty module is proposed to measure the point-wise uncertainty by hierarchically  increasing contextual ranges, which enables progressively capturing the local context at different scales.
Then, the neighbouring points with similar features are removed by the feature distance suppression module, thereby further reducing the annotation redundancy. 
The remaining points with high uncertainty scores as well as sparse distributions are regarded as the candidates to be annotated. In this manner, the most valuable and representative unlabelled points are selected to expand the labelled set, with which the semi-supervised point cloud semantic segmentation neural network can be further enhanced. In the following, we describe each component of the proposed framework in detail.

\subsection{Hierarchical Minimum Margin Uncertainty Measurement}
 In the proposed method, we aim to select and label the most valuable points that can bring maximum performance improvements. Intuitively, identifying and using the most informative labelled data for training is the key to obtaining a model with good quality, and consequently accurate segmentation prediction. 
Merely considering individual points without their surrounding context information cannot reflect the real importance of the point. 
Thus, we design the hierarchical minimum margin uncertainty measurement module to calculate the uncertainty score for each point by grouping points at multiple scales and progressively perceiving context information in a broader range along the hierarchy for the unlabelled point.

As shown in Figure~\ref{fig:framework}, the prediction for each point generated with the currently trained model is input into the HMMU module. 
 We first calculate the point-level minimum margin uncertainty score $U_{x}$ for each unlabelled point with Eq. \ref{eq:point uncertainty}:
\begin{equation}
    \begin{aligned}
    \centering
    U_{x} &= h(x^u; p_1(x^u)) - h(x^u; p_2(x^u)),
    \end{aligned}
	\label{eq:point uncertainty}
\end{equation}
where $x^u$ is the candidate point that is chosen to be labelled; $p_1(x^u)$ and $p_2(x^u)$ are the highest and second-highest prediction of $x^u$ under the segmentation predictor $h(\cdot)$.


Then we group a wider range of neighbours through downsampling. The softmax prediction of each downsampled point is obtained by averaging the softmax prediction of its neighbouring points in the original point cloud. Each softmax label of the downsampled point represents the prediction distribution of a local region, which is denoted by ${S_R}$.

We perform $N$ levels of voxel downsampling. The local context information of the to-be-annotated point $x^u$ at the $i_{th}$ level of downsampling, denoted by $S_R^i(x^u)$, can be represented as the average of predictions of the grouped points in the following:
\begin{equation}
    \centering
    S_R^i(x^u) = \frac{1}{K}\underset{j=1}{\overset{K}{\sum}}p(x_j^u),
\end{equation}
where $x_j^u$ is the $j_{th}$ neighbour for $x^u$ in the original point cloud; $K$ is the number of neighbours within the predefined  voxel radius $v_r$; and $p(\cdot)$ is the prediction probability. 


 
For each level of downsampling,  the voxel-level contextual uncertainty scores $U_R^i$ for the unlabelled point $x^u$ can be obtained with:
\begin{equation}
    \begin{aligned}
    \centering
	U_{R}^i &= h(x^u; S_{R1}^i(x^u)) - h(x^u; S_{R2}^i(x^u)),
    \end{aligned}
	\label{eq:Region uncertainty}
\end{equation}
where $S_{R1}^i(x^u))$ and $S_{R2}^i(x^u)$  are the highest and the second highest average prediction of the grouped points of $x^u$ from the  segmentor $h(\cdot)$ in the $i_{th}$ downsampling. 

 Finally, for each unannotated point $x^u$, we integrate the point-level and  the voxel-level contextual uncertainty scores to get the final uncertainty  $v^u$ using Eq. \ref{eq:final uncertainty}:
\begin{equation}
    \begin{aligned}
    \centering
	v^u &= U_x + \underset{i=1}{\overset{N}{\sum}} \omega^i \times U_{R}^i.
    \end{aligned}
	\label{eq:final uncertainty}
\end{equation}
Here, $\omega^i \in \{0.1, 0.01, 0.001\}$ is the hyperparameter representing the fusion weight of the contextual uncertainty at the $i_{th}$ level of downsampling and three levels of downsampling is performed in the experiment.

With HMMU, the context information is  fused to significantly improve performance by merely annotating points rather than regions.

\subsection{Feature-distance Suppression based Point Selection}

 \begin{figure}[t]
    \begin{center}
   \includegraphics[width=0.9\linewidth]{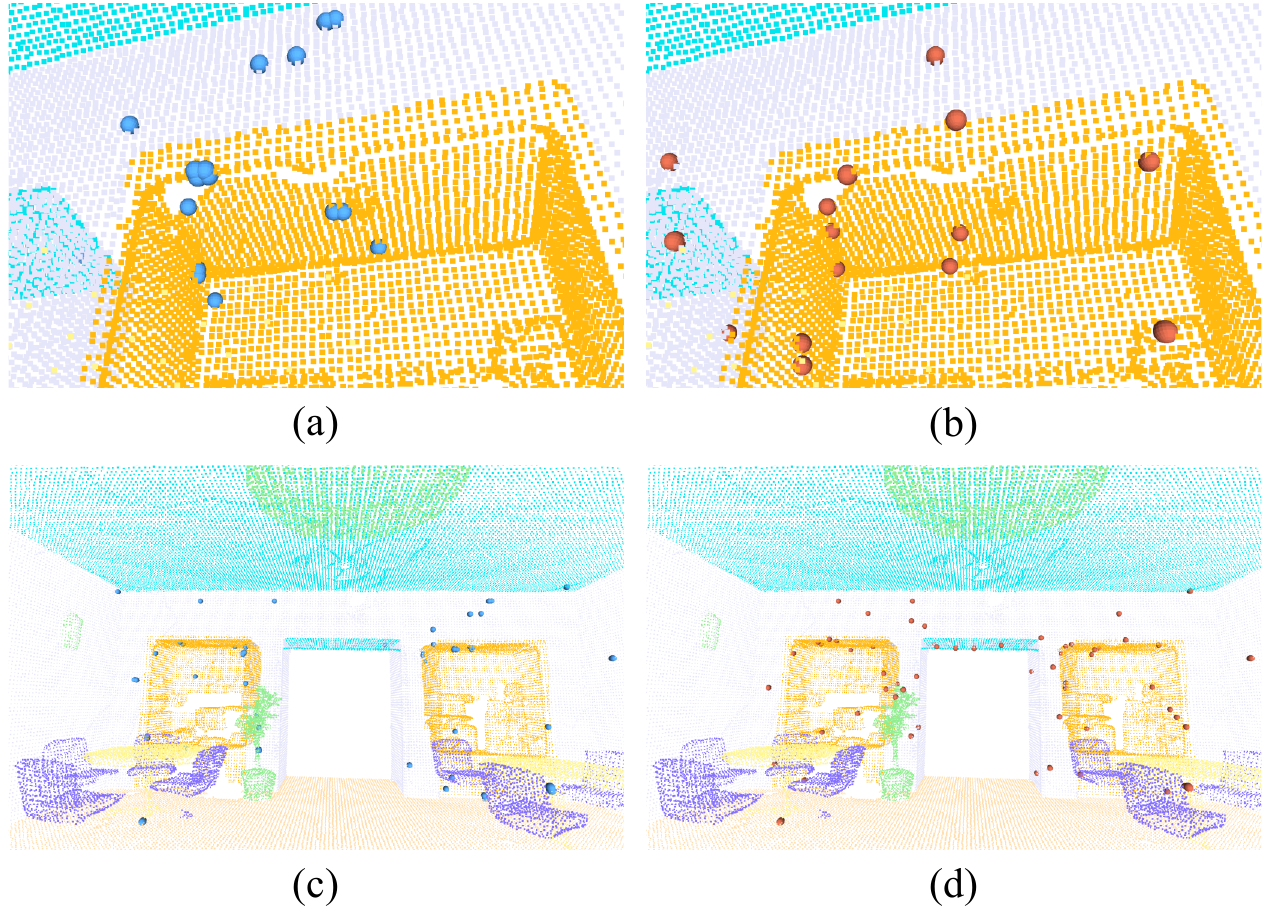}
    \end{center}
    \vspace{-1.5em}
   \caption{Visualization of selected points. (a) and (c) show the points selected using the Top-$K$ approach (blue dots); (b) and (d) present points chosen by FDS (orange dots). FDS enables a more spread-out point selection.}
   \vspace{-1em}
\label{fig:scannetvis}
\end{figure}
With the point-wise uncertainty score $v^u$, we can select the Top-$K$ points with the highest uncertainty to label. However, as shown in Figure \ref{fig:scannetvis}, such a point selection way might make points concentrate in local areas, causing label redundancy. Thus, we propose a feature-distance suppression (FDS) module to ensure the selected points retain a spread-out distribution in the space, thus offering a more effective overall representation. 

Given a distance suppression radius $r$ and a feature similarity threshold $\tau$, for a point $x_i$ to be selected, we first determine whether there are points within its radius $r$ that have been already selected. 
 A judgment set $D^i$ is constructed for the candidate point $x_i$ in the current point selection iteration.
\begin{equation}
     \begin{aligned}
    \centering
	D^i &= D^i \cup \{x_j\}, if \\
        \forall{x_j} & \hspace{0.1cm}is \hspace{0.1cm} selected:
        {d}_{ij} < r,
    \end{aligned}
\end{equation}
where $x_j$ is the neighbour of $x_i$ within the radius of $r$; $d_{ij}$ is the Euclidean distance between $x_i$ and $x_j$; and $D^i$ represents if there are other candidate points in the neighbourhood and $D^i$ is initialized to be empty.

If there is no neighbour of $x_i$ that is selected as a candidate, i.e. the judgment set $D^i$ is empty, $x_i$ is regarded as the valuable point that can represent the local area for labelling. Conversely, if the neighbouring point $x_j$ is already selected as the uncertainty point, cosine similarity $Sim_{ij}$ between the features of $x_i$ and $x_j$ is then computed with Eq. \ref{eq:sim}. 
\begin{equation}
    \begin{aligned}
    \centering
	Sim(x_i,x_j) &= \frac{\mathbf{f}_i \cdot \mathbf{f}_j}{||\mathbf{f}_i||\cdot||\mathbf{f}_j||},
    \end{aligned}
	\label{eq:sim}
\end{equation}
where $\mathbf{f}_i$ and $\mathbf{f}_j$ are the features of $x_i$ and $x_j$ extracted with the trained student segmentation network.

If there exists an ${x_j}$ in $D^i$ for which $Sim_{ij} > \tau$, $x_i$ is regarded as a redundant point with respect to the labelled data and $x_i$ is excluded from labelling.


\subsection{Training objective}
 To train a network with partly annotated data, a usual approach is to apply the cross-entropy loss to the labelled points and ignore the unlabelled ones, which causes inadequate supervision for model learning. Inspired by previous semi-supervised 3D point cloud segmentation methods, in the proposed method we employ a teacher-student framework to exploit unlabelled points and provide additional supervision.
Specifically, 
we construct two segmentation networks using MinkowskiNet~\cite{choy20194d}, denoted as the teacher model ${M}_t$ and student model ${M}_s$, respectively. A consistency constraint between the teacher model and the student model is exploited to learn knowledge from unlabelled data. Pseudo-labels of unlabelled data are generated using the teacher model. During training, the original sample is input into the teacher network while its augmented counterparts are input into the student. 
For those labelled points, we apply the standard cross-entropy loss. For the unlabelled ones, we generate pseudo-labels from the output of the teacher and also compute the cross-entropy loss. The losses in both labelled and unlabelled points are then used to optimize the student network by gradient descent. After updating the student model at each step, Exponential Moving Average (EMA) is used to transfer the parameters of the student model to the teacher model:
 \begin{equation}
    \begin{aligned}
    \centering
	 \theta_t^j = \alpha\theta_t^{j-1} + (1-\alpha)\theta_s^j,
    \end{aligned}
	\label{eq:loss_unsup}
\end{equation}
where $\theta_t$ and $\theta_s$ are the parameters of the teacher and student networks, respectively, and $j$ denotes the $j$-$th$ training
step. $\alpha$ is the hyper-parameter to determine the speed of
parameter transmission, which is generally close to 1.

\section{Experiments}
\subsection{Datasets}
We evaluate the proposed approach on two publicly available benchmark datasets, S3DIS \cite{armeni20163d} and ScanNetV2 \cite{dai2017scannet}, to demonstrate its superior performance.

\textbf{S3DIS} (Stanford Large-Scale 3D Indoor Space Dataset) is a large-scale indoor point cloud dataset captured by Matter-port scanners which mainly dedicate to 3D segmentation tasks. It consists of 6 different large-scale indoor areas with a total of 271 rooms, each room is a separate point cloud sample that includes coordinate, colour, and annotation information for each point. Following the standard evaluation criterion, Area 5 is used as the test set and the rest of the areas are used as the training set.

\textbf{ScanNetV2} is an RGB-D video dataset. It contains 1513 point cloud samples from 707 individual indoor scenes for training and 100 unlabelled samples for testing. For the 3D semantic segmentation task, ScanNetV2 provides up to 40 categories of classification labels and selects 20 categories as classification targets to establish the segmentation task. Each point in the dataset contains coordinate, colour, and annotation information. In the experiments, we follow the official data split, dividing the 1513 training data into 1201 training samples and 312 validation samples, and evaluate methods on the test set.
\subsection{Implementation Details}
\textbf{Segmentation Model.}
In our experiments, we use the PyTorch implementation of MinkowskiNet in ReDAL \cite{wu2021redal} as our backbone network with the stochastic gradient descent (SGD) optimizer. The architecture is trained on a single NVIDIA RTX A6000 GPU. 

\textbf{Training Settings. }
For S3DIS, we set the batch size to 4, and train 60K steps per iteration. The initial learning rate is 0.1, and cosine annealing with a period of 5000 is used as the learning rate decay strategy. For ScanNetV2, we set the batch size to 4, and train 30K steps per iteration. The initial learning rate is 0.1, and the learning rate decay strategy is using polynomial decay with a power of 0.9. In the teacher-student model, 
the keep rate of EMA and the thresholds for the pseudo-label are set to $ 0.955$  and $0.75$, respectively.  $5$ iterations are performed in the active learning framework. In HMMU, the point cloud is downsampled at each scale with the sampling radius of $10$cm, $50$cm, and $100$cm, respectively. In the FDS module, the  radius of the local region is set to $20$cm and the feature similarity threshold is set to $0.8$.

\subsection{Comparison with Active Learning Methods}
To demonstrate the effectiveness of our active learning strategy, we remove the semi-supervised module in the comparisons with active learning methods. As  current active learning methods for point cloud semantic segmentation are region-based, we compare our proposed method against them to verify the effectiveness of the point-based active learning method. 
The compared results are shown in Table \ref{tab: active learning comparison}.  Compared with the region-based active learning methods, our point-based selection strategy greatly reduces the labelling cost, with the demand of achieving $90\%$  performance of the fully-supervised baselines. With only $0.1\%$ (about 13,000 points) of the downsampled points, our method allows for $90\%$ of the performance of our fully supervised baseline, i.e. fully-supervised MinkowskiNet.
This verifies our theory that the strong semantic similarity in local areas of point clouds results in redundant labelling and ineffective training in region-based active learning.
\begin{table}[h]
    \centering
    \resizebox{0.9\linewidth}{!}{
    \begin{tabular}{llc}
        \hline
		Methods & Segmentors & Labelled points\\
		\hline\hline
        ReDAL \cite{wu2021redal} & SPVCNN & 13\%\\
		ReDAL \cite{wu2021redal}  & MinkowskiNet & 15\%\\
        SSDR-AL \cite{shao2022active} & Randlanet & 11.7\%\\
        \hline
        Ours & MinkowskiNet & 0.1\%\\
		\hline		
    \end{tabular}
    }
    \caption{The comparison of the percentage of labelled points required to achieve 90\% accuracy of fully-supervised baseline on the S3DIS dataset against  current region-based active learning methods. For a fair comparison, the semi-supervised module of our method is removed here which means only labelled data and selected points are used during training.
    }
    \vspace{-0.8em}
    \label{tab: active learning comparison}
\end{table}

Furthermore, we compare the performance of our active learning strategy with existing region-based ones and four basic point-based selection strategies including random selection (Random), softmax least confidence (LC), softmax minimum margin uncertainty (MMU) and softmax entropy (Entropy). The results are shown in Table~\ref{tab: comp_pal}. It can be seen that the performance of our method is increased by $4.6\%$ compared with random point  selection and surpasses the state-of-the-art region-based active learning method by $1\%$ with significantly fewer labelled points ($0.1\%$ v.s. $11.7\%$).

\begin{table}[h]
    \centering
    \resizebox{\linewidth}{!}{
    \begin{tabular}{lllcc}
        \hline
		Type & Methods & Segmentors & labelled points & mIoU(\%)\\
		\hline\hline
        \multirow{3}*{Region-based} & ReDAL & SPVCNN & 15\% & 58.0\\
         & ReDAL & MinkowskiNet & 15\% & 57.3\\
         & SSDR-AL & Randlanet & 11.7\% & \textbf{58.3}\\
        \hline
        \multirow{5}*{Point-based} & Random & MinkowskiNet & 0.1\% & 54.7\\
        & Entropy & MinkowskiNet & 0.1\% & 48.2\\
         & LC & MinkowskiNet & 0.1\% & 53.3 \\
         & MMU & MinkowskiNet & 0.1\% & 55.5 \\
        \cline{2-5}
        & Ours & MinkowskiNet & 0.1\% & \textbf{59.3}\\
		\hline		
    \end{tabular}}
    \caption{The comparison of mIoU between our active strategy without the semi-supervised module and different active learning methods on the S3DIS dataset.}
    \label{tab: comp_pal}
\end{table}

\subsection{Comparison with Weakly-supervised Methods}  
We also compare the proposed method with the state-of-the-art weakly-supervised methods. Similar to \cite{li2022hybridcr,yang2022mil}, we   present state-of-the-art segmentation methods with different supervision settings. The comparison results on the datasets of S3DIS and ScannetV2 are presented in Table \ref{tab:weakly-supervised methods on S3DIS dataset} and Table \ref{tab:weakly-supervised methods on ScanNet dataset}, respectively.


\begin{table}[h]
    \scriptsize
    \centering
    \resizebox{0.9\linewidth}{!}{
    \begin{tabular}{c|c|c}
        \hline
		Methods & Labelled points & Area 5\\
		\hline\hline
        PointNet \cite{qi2017pointnet} & 100\% & 41.1\\
        PointCNN \cite{li2018pointcnn}& 100\% & 57.3\\
        KPConv \cite{thomas2019kpconv}& 100\% & 67.1\\
        PointTransformer \cite{zhao2021point} & 100\% & 70.4\\
        CBL \cite{tang2022contrastive} & 100\% & 69.4\\
        MinkowskiNet$^\dagger$ \cite{choy20194d} & 100\% & 64.5\\
        \hline
        PSD \cite{zhang2021perturbed}& 1\% & 63.5\\
        SQN \cite{hu2022sqn} & 1\% & 63.7\\
        HybridCR \cite{li2022hybridcr}& 1\% & 65.3\\
        Ours & 0.43\% & \textbf{65.7}\\
        \hline
        SQN \cite{hu2022sqn} & 0.1\% & 61.4\\
        Ours & 0.07\% & \textbf{62.3}\\
		\hline
        PSD \cite{zhang2021perturbed}& 0.03\% & 48.2\\
        HybridCR \cite{li2022hybridcr}& 0.03\% & 51.5\\
        MIL \cite{yang2022mil} & 0.02\% & 51.4\\
        GaIA \cite{lee2023gaia}& 0.02\% & 53.7\\
        VIB \cite{tian2022vibus} & 0.02\% & 52.0\\
        Ours & 0.02\% & \textbf{55.9}\\
        \hline
    \end{tabular}
    }
    \caption{Comparison with existing weakly-supervised methods on S3DIS Area-5. $^\dagger$ represents the result of the baseline trained on our own device.}
    \label{tab:weakly-supervised methods on S3DIS dataset}
    \vspace{-1em}
\end{table}

\textbf{S3DIS.}
we perform experiments under three different annotation budgets, namely 0.7\%, 0.07\%, and 0.02\%. For the sake of fairness, we follow the SQN \cite{hu2022sqn} to calculate the annotation percentage which is defined as the ratio between the number of labelled data and all training and validation data. Each annotation budget is completed through five iterations.
As shown in Table 3, our approach gains consistent improvements in the segmentation performance with labelled points increased from $0.02\%$ to $0.43\%$. It is worth noting that, for the $0.7\%$ annotation budget, the segmentation performance at the third iteration ($0.43\%$ annotations) has already surpassed the fully-supervised MinkowskiNet.
This verifies that selecting the most representative points and making full use of the remaining unlabelled data can greatly benefit the performance of point cloud semantic segmentation.  Our method also outperforms the state-of-the-art PSD, SQN and HybridCR under $1\%$ labelled data by $2.2\%$, $2.0\%$ and $0.4\%$, respectively. When the number of labelled points is reduced to $0.07\%$, our method is also superior to the SQN with $0.1\%$ labelled points and achieves $96.5\%$ performance of the fully-supervised baseline (MinkowskiNet).  In the case of the least labelling budget,
we evaluate our method in the setting of $0.02\%$ labelled points. 
It can be seen that our method also achieves the best performance, outperforming the state-of-the-art GaIA, MIL, HybridCR and PSD by $2.2\%$, $4.5\%$, $4.4\%$ and $7.7\%$. 

\begin{figure*}[h]
    \begin{center}
   \includegraphics[width=1\linewidth]{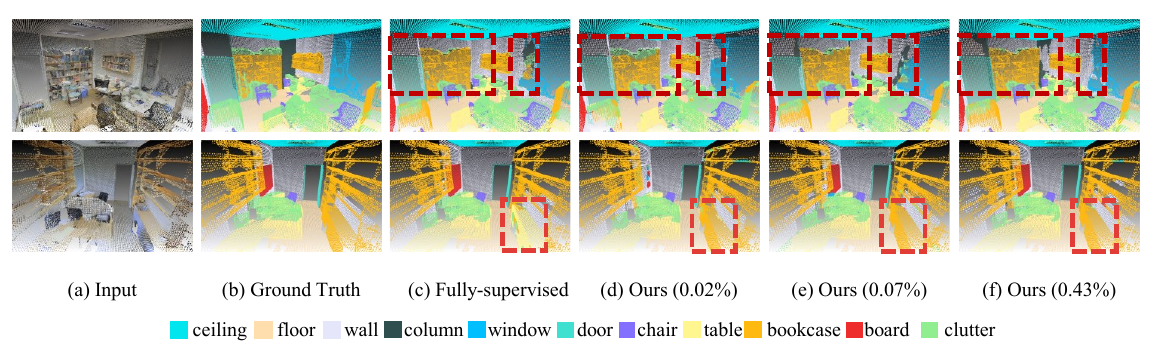}
    \end{center}
    \vspace{-1em}
   \caption{Visualization of segmentation results on the test set of S3DIS Area-5. Our method achieves comparable or even better results than our fully-supervised baseline (MinkowskiNet) when 0.43\% of the labelled data is adopted.}
\label{fig:viscomp1_s3dis}
\end{figure*}

We also visualize the segmentation results of our method under different annotation budgets in Figure \ref{fig:viscomp1_s3dis}. The results show that with $0.43\%$ labelled data, our approach can achieve comparable segmentation results with the fully supervised baseline. Benefiting from the points selected by our method, our segmentation results are even more promising in some cases, such as the bookcase and column.

\begin{figure*}[t]
    \begin{center}
   \includegraphics[width=1\linewidth]{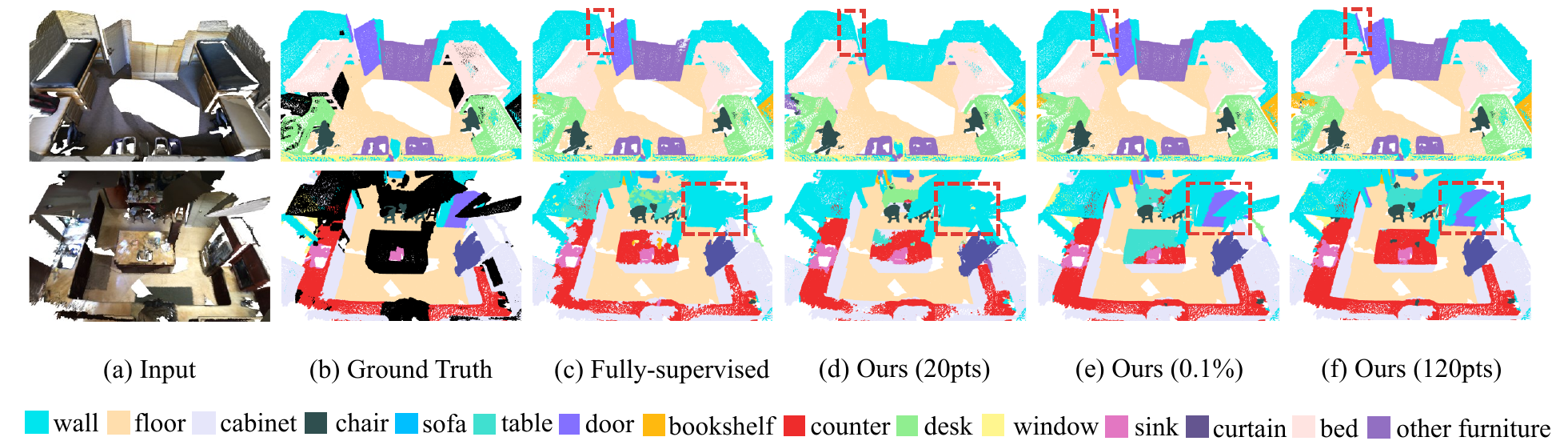}
    \end{center}
   \caption{Visualization of segmentation results on the validation set of ScannNetV2. Our method achieves comparable or even better results
than our fully-supervised baseline (MinkowskiNet) when 120 labelled points per scene are adopted.}
\vspace{-1em}
\label{fig:Scannetvis2}
\end{figure*}
\begin{table}[h]\scriptsize
    \centering
    \resizebox{0.9\linewidth}{!}{
    \begin{tabular}{c|c|cc}
        \hline
	Methods & Labelled points & val & test\\
	\hline\hline
        PointNet++ \cite{qi2017pointnet++} & 100\% & - & 33.9\\
        KPConv  \cite{thomas2019kpconv}& 100\% & - & 68.4\\
        VMNet \cite{hu2021vmnet} & 100\% & - & 74.6\\
        BPNet \cite{hu2021bidirectional} & 100\% & - & 74.9\\
        MinkowskiNet$^\dagger$ \cite{choy20194d}& 100\% & 69.3 & 68.0\\
        \hline
        PSD \cite{zhang2021perturbed} & 1\% & - & 54.7\\
        HybridCR \cite{li2022hybridcr} & 1\% & 56.9 & 56.8\\
        GaIA \cite{lee2023gaia} & 1\% & - & 65.2\\
        \hline
        SQN \cite{hu2022sqn} & 0.1\% & - & 56.9\\
        Ours & 0.1\% & 69.9 & \textbf{68.2}\\
        \hline
        SQN \cite{hu2022sqn} & 200pts & - & 59.8\\
        CSC \cite{hou2021exploring} & 200pts & 68.2 & 66.5\\
        GaIA \cite{lee2023gaia} & 200pts & - & 68.5\\
        VIBUS \cite{tian2022vibus} & 200pts & 69.6 & 69.1\\
        Ours & 120pts & 70.2 & \textbf{69.4}\\
        \hline
        SQN \cite{hu2022sqn} & 20pts & - & 48.6\\
        CSC \cite{hou2021exploring} & 20pts & 55.5 & 53.1\\
        MIL \cite{yang2022mil} & 20pts & 57.8 & 54.4\\
        VIBUS \cite{tian2022vibus} & 20pts & - & 58.6\\
        GaIA \cite{lee2023gaia}& 20pts & - & \textbf{63.8}\\
        Ours & 20pts & 62.2 & 62.5\\
        \hline
    \end{tabular}
    }
    \caption{The comparison with existing weakly-supervised methods. $^\dagger$ is the result of the baseline trained on our device.}
    \vspace{-2.5em}
    \label{tab:weakly-supervised methods on ScanNet dataset}
\end{table}
\textbf{ScanNetV2.}
As illustrated in Table~\ref{tab:weakly-supervised methods on ScanNet dataset}, we use the criteria of percentage and fixed number of labelled points as different annotation budgets to evaluate the methods, including 0.1\%, 200pts, and 20pts. Similar to S3DIS, each annotation budget completes through five iterations, and the third iteration (120pts) is reported in the case of 200pts. We substantially beat SQN's performance by $11.3\%$ using $0.1\%$ of annotation. Compared against PSD, HybridCR and GaIA which are trained with $1\%$ annotations, our method outperforms them with only $0.1\%$ labels by $13.5\%$, $11.4\%$ and $3\%$, respectively. 
In the case of the fixed number of annotation points, we use 200 points per scene (200pts) and 20 points per scene (20pts) as the annotation budget. In our experiment with 200pts, when the labelling points reach 120pts, our method has achieved comparable performance to the fully supervised baseline. This proves the effectiveness of our active learning strategy on the semi-supervised point cloud semantic segmentation.
We also evaluate our method on the least annotation setting. In the case of 20pts, our result is much higher than SQN, CSC, MIL, and VIBUS, showing that our method effectively selects the most valuable points. However, our result is slightly lower than GaIA, as the performance of our feature extraction backbone is limited given only 20pts (the proportion is less than $0.02\%$ of ScanNetV2).

As shown in Fig. \ref{fig:Scannetvis2}, we visualize the segmentation results of our method under different annotation budgets on  ScanNetV2. 
It is observed that our method can achieve comparable or even better segmentation results with minor training budgets compared to our fully-supervised baseline. 
\begin{figure}[h]
    \begin{center}
   \includegraphics[width=0.9\linewidth]{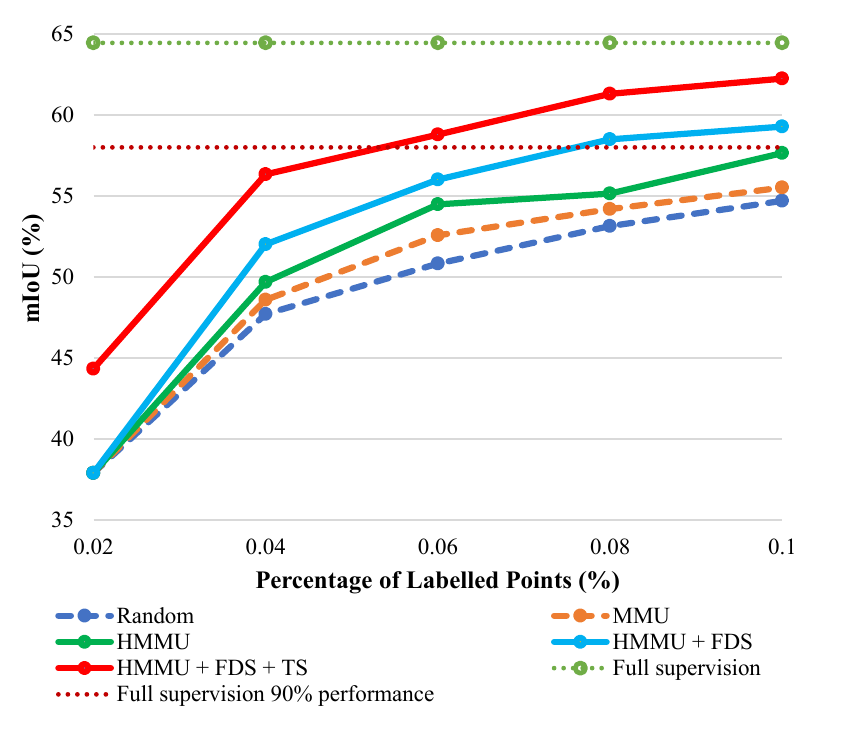}
    \end{center}
    \vspace{-1.5em}
   \caption{The comparison of performance improvements for different combinations of modules.}
   \vspace{-1em}
\label{fig:ablation_study}
\end{figure}


\subsection{Ablation Studies}
To demonstrate the effectiveness of each module in our method, we conduct the following ablation studies on S3DIS dataset. Following \cite{hu2022sqn}, all the ablated networks are trained using 0.1\% labelled points on Areas 1, 2, 3, 4, 6 and tested on Area 5. The number of iterations is set to 5 and 0.02\% data is selected in each iteration. To clearly see the effect of active learning, we present the results as shown in Figure \ref{fig:ablation_study}. 
We take random sampling as the basic strategy for active selection where points are randomly selected in each iteration. After we replace the random sampling with MMU for active sampling, the mIoU is slightly improved by $1.5\%$. After utilising HMMU to select points, we acquire an improvement of around $3.0\%$ compared with MMU in the last iteration. 
The FDS module is also added to further improve the performance. HMMU+FDS achieves an extra $2.9\%$ mIoU improvement compared to HMMU. More impressively, merely relying on the active learning modules including HMMU and FDS, our approach already outperforms the random sampling baseline by $8\%$ and surpasses $90\%$ performance of full-supervision with only $0.1\%$ labelling data.
The effectiveness of the semi-supervised module is also demonstrated by the experimental results of HMMU+FDS+TS. Obviously, adding the semi-supervised teacher-student module, the segmentation result is further improved  by $5\%$ compared to HMMU + FDS. The proposed method also reaches $95\%$ of the fully-supervised setting with only $0.1\%$ labelled data.


In Table \ref{tab: ablation2}, we show the effectiveness of different modules by quantitative comparison.
We compare different compositions of our module to verify their effectiveness. Shown in the first row, for the base case where unlabelled points are selected by MMU in each iteration, the mIoU is only $55.51\%$. After using the HMMU module, the segmentation results are improved to $57.65\%$. If we only apply FDS to the MMU selection for each iteration, a slight improvement of $1.17\%$ is obtained compared with the based case. When the active learning strategy is removed, as shown in the fourth row in Table \ref{tab: ablation2}, the semi-supervised segmentation method only achieves the performance of $57.01\%$. Point cloud semantic segmentation only using the active learning technique can reach the mIoU of $59.29\%$. If we combine semi-supervised and active learning, as shown in the last row, the performance can be improved to  $62.26\%$.
\begin{table}[h]\scriptsize
    \centering
    \resizebox{0.3\textwidth}{!}{
    \begin{tabular}{ccc|c}
        \hline
        \multicolumn{3}{c|}{Components} & \multicolumn{1}{c}{\multirow{2}*{mIoU(\%)}}\\
        \cline{1-3}
		HMMU & FDS & TS & \\
		\hline\hline
       base. &  & & 55.51\\
       \checkmark & & & 57.65\\
        & \checkmark & & 56.68 \\
        &  & \checkmark & 57.01 \\
       \checkmark & \checkmark & & 59.29\\
       \checkmark & \checkmark & \checkmark & \textbf{62.26}\\
		\hline		
    \end{tabular}}
    \caption{Ablation study of different components.}
    \label{tab: ablation2}
    \vspace{-1em}
\end{table}


In Table \ref{tab:HMMU_layer}, we show the ablation study of each layer in the HMMU module. PL represents the  layer that calculates the point-level minimum margin
uncertainty score. VL is the layer which calculates the voxel-level contextual uncertainty score.  
$v_r$ and $\omega$ are the voxel-radius and weight for each VL layer.
It is shown that the combination of one layer of point-level uncertainty and three layers of voxel-level uncertainty can achieve the best performance. 
We also show the ablation study on the hyperparameters of the FDS module in Table \ref{tab:t4}. We achieve the best segmentation result given that $r$ and $\tau$ are set to $20$cm and $0.8$ respectively. 
Thus, we adopt this combination in our experiments.

\begin{table}[h]
\centering
	\resizebox{0.48\textwidth}{!}
   {
		\begin{tabular}{l|ll|ccccc}
                \hline
                 \multirow{2}*{Layer}&\multicolumn{2}{c|}{HMMU} & \multirow{2}*{iter 1} & \multirow{2}*{iter 2} & \multirow{2}*{iter 3} & \multirow{2}*{iter 4} & \multirow{2}*{iter 5}\\
			&$\omega$ & $v_r$ (cm)\\
			\hline
                \hline
                PL& -- & -- & 37.83 & 50.37 & 54.32 & 56.18 & 56.68 \\
                PL + VL1&\{0.1\} & \{10\} & 37.83 & 52.71 & 54.84 & 56.52 & 57.03 \\
                PL + VL2&\{0.01\} & \{50\} & 37.83 & 51.45 & 55.81 & 57.22 & 57.43 \\
                PL + VL3&\{0.001\} & \{100\} & 37.83 & 51.06 & 55.62 & 57.36 & 58.03 \\
                PL + VL1,2&\{0.1, 0.01\} & \{10, 50\} & 37.83 & 51.56 & 55.28 & 56.66 & 58.21 \\
                PL + VL1,2,3&\{0.1, 0.01, 0.001\} & \{10, 50, 100\} & 37.83 & 52.02 & 56.02 & 58.49 & \textbf{59.29} \\
                PL + VL1,2,3,4&\{0.1, 0.01, 0.001, 0.001\} & \{10, 50, 100, 200\} & 37.83 & 51.56 & 56.32 & 57.61 & 57.71  \\
                PL + VL1,2,3,4&\{0.1, 0.01, 0.001, 0.0001\} & \{10, 50, 100, 200\} & 37.83 & 51.69 & 56.22 & 58.53 & 59.19 \\
                \hline
	   \end{tabular}
    }
     \caption{Ablation study of HMMU layers.}
     \vspace{-1.0em}
	\label{tab:HMMU_layer}
\end{table}

\begin{table}[h]
	\centering
	\resizebox{0.48\textwidth}{!}{
		\begin{tabular}{cc|ccccc}
                \hline
                \multicolumn{2}{c|}{FDS} & \multirow{2}*{iteration1} & \multirow{2}*{iteration2} & \multirow{2}*{iteration3} & \multirow{2}*{iteration4} & \multirow{2}*{iteration5}\\
			r(cm) & $\tau$\\
			\hline
                \hline
                20 & 0.8 & 37.83 & \textbf{52.02} & \textbf{56.02} & \textbf{58.49} & \textbf{59.29} \\
                20 & 0.9 & 37.83 & 51.27 & 54.90 & 56.79 & 58.79\\
                40 & 0.8 & 37.83 & 51.35 & 55.71 & 57.09 & 57.79 \\
                40 & 0.9 & 37.83 & 51.65 & 55.26 & 57.45 & 59.13 \\
                50 & 0.95 & 37.83 & 50.87 & 55.73 & 56.49 & 57.82\\
                \hline
	   \end{tabular}
    }
        \caption{Ablation study of the hyperparameters in FDS module.}
        \vspace{-1em}
	\label{tab:t4}
\end{table}

\section{Conclusion}

In this paper, we study active learning for 3D point cloud semantic segmentation. In comparison with previous region-based methods, we propose a hierarchical point-based active learning strategy consisting of two new designs, HMMU and FDS. HMMU serves as an effective way to measure the uncertainty or importance of labelling, while FDS enables us to select the most valuable points by considering the feature similarity and spatial distribution. To better use the unlabelled points during network learning, we also introduce a teacher-student structure to generate pseudo-labels.  Extensive experiments are conducted on S3DIS and ScanNet datasets to demonstrate the effectiveness of our method. In the future, a more advanced semi-supervised architecture can be adopted for generating more reliable pseudo-labels to further facilitate active learning based point cloud semantic segmentation tasks.

\textbf{Acknowledgements}. This work is supported by the National Natural Science
Foundation of China (No. 62206033, 
 No.62036007, No.62221005), and the Natural Science Foundation of Chongqing (No. cstc2020jcyj-msxmX0855, No. cstc2021ycjh-bgzxm0339).

{\small
\bibliographystyle{ieee_fullname}
\bibliography{egbib}
}

\end{document}